%% file: PaperForReview.tex
\documentclass[10pt,twocolumn,letterpaper]{article}

\usepackage{cvpr}              

\usepackage{caption}
\usepackage[labelformat=simple]{subcaption}

\usepackage{times}
\usepackage{epsfig}
\usepackage{graphicx}
\usepackage{amsmath}
\usepackage{amssymb}
\usepackage{amsfonts}
\usepackage{enumitem}
\usepackage{kotex}
\usepackage{booktabs}
\usepackage{tabularx}
\usepackage{multirow}
\usepackage{arydshln}
\usepackage{algorithm}
\usepackage{algpseudocode}
\usepackage{url}

\usepackage{bm}
\usepackage{subcaption}
\usepackage{stfloats}
\usepackage{arydshln}
\usepackage{adjustbox}

\usepackage[pagebackref,breaklinks,colorlinks]{hyperref}

\usepackage[capitalize]{cleveref}
\crefname{section}{Sec.}{Secs.}
\Crefname{section}{Section}{Sections}
\Crefname{table}{Table}{Tables}
\crefname{table}{Tab.}{Tabs.}

\def\cvprPaperID{*****} 
\def\confName{CVPR}
\def\confYear{2022}

\begin{document}

\title{Vision Transformer for Small-Size Datasets}

\author{Seung Hoon Lee\\
Inha University\\
Incheon, South Korea\\
{\tt\small aanna0701@gmail.com }
\and
Seunghyun Lee\\
Inha University\\
Incheon, South Korea\\
{\tt\small lsh910703@gmail.com}
\and
Byung Cheol Song\\
Inha University\\
Incheon, South Korea\\
{\tt\small bcsong@inha.ac.kr}
}
\maketitle

\begin{abstract}
Recently, the Vision Transformer (ViT), which applied the transformer structure to the image classification task, has outperformed convolutional neural networks. However, the high performance of the ViT results from pre-training using a large-size dataset such as JFT-300M, and its dependence on a large dataset is interpreted as due to low locality inductive bias. This paper proposes Shifted Patch Tokenization (SPT) and Locality Self-Attention (LSA), which effectively solve the lack of locality inductive bias and enable it to learn from scratch even on small-size datasets. Moreover, SPT and LSA are generic and effective add-on modules that are easily applicable to various ViTs. Experimental results show that when both SPT and LSA were applied to the ViTs, the performance improved by an average of $2.96\%$ in Tiny-ImageNet, which is a representative small-size dataset. Especially, Swin Transformer achieved an overwhelming performance improvement of $4.08\%$ thanks to the proposed SPT and LSA.
\end{abstract}

\section{INTRODUCTION}
\label{sec1:intro}
\input{figures/FigMain}
    Convolutional neural networks (CNNs), which are effective in learning visual representations of image data, have been the main-stream in the field of computer vision (CV)~\cite{conv:dense,conv:efficient,conv:google,conv:res,conv:alex,conv:HRNet}. Meanwhile, in the field of Natural Language Processing (NLP), the so-called Transformer~\cite{AttentionIA} based on self-attention mechanism has achieved tremendous success~\cite{Bert, nlp:albert, nlp:gpt3}. So, in the CV field, there have been attempts to combine the self-attention mechanism with CNNs~\cite{attention:NLNN, attention:AttentionAugmented, attention:ResidualAN, attention:slef-attention, attnetion:squeeze, attention:BoT}. These studies have succeeded in proving that the self-attention mechanism also works for the image domain. Recently, it was reported that Vision Transformer (ViT)~\cite{vit:vit}, which applied a standard Transformer composed entirely of self-attention to image data, showed better performance than ResNet~\cite{conv:res} and EfficientNet~\cite{conv:efficient} in the image classification task. This made Transformer receive a lot of attention in the CV field.

    
    ViT rarely uses convolutional filters, i.e., the core of CNNs. Convolutional filters were usually used only for their tokenization. Thus, ViT structurally lacks locality inductive bias than CNNs, and they require a too large amount of training data to obtain acceptable visual representation~\cite{inductive2}. For example, just to learn a small-size dataset, ViT had to precede pre-training on a large-size dataset such as JFT-300M~\cite{dataset:JFT}. In order to alleviate the burden of pre-training, several ViTs which can learn a mid-size dataset such as ImageNet from scratch have been proposed. Such data-efficient ViTs tried to increase the locality inductive bias in terms of network architecture. For example, some adopted a hierarchical structure like CNNs to leverage various receptive fields~\cite{vit:swin, vit:pit, vit:CvT}, and the others tried to modify the self-attention mechanism itself~\cite{vit:CaiT, vit:Lovit, vit:swin, vit:CvT, vit:Cos}. However, learning from scratch on mid-size datasets still requires significant costs. Moreover, learning small-size datasets from scratch is very challenging considering the trade-off between dataset capacity and performance. Therefore, we need to study ViT that can learn small-size datasets by sufficiently increasing the locality inductive bias.
    
    
    Through observations, we found two problems that decrease locality inductive bias and limit the performance of the ViT. The first problem is poor tokenization. ViT divides a given image into non-overlapping patches of equal size, and linearly projects each patch to a visual token. Here, the same linear projection is applied to each patch. So, tokenization of the ViT has the permutation invariant property, which enables a good embedding of relations between patches~\cite{inductivebias}. On the other hand, non-overlapping patches allow visual tokens to have a relatively small receptive field. Usually, tokenization based on non-overlapping patches has a smaller receptive field than tokenization based on overlapping patches with the same down-sampling ratio. Small receptive fields cause ViT to tokenize with too few pixels. As a result, the spatial relationship with adjacent pixels is not sufficiently embedded in each visual token. The second problem is the poor attention mechanism. The feature dimension of image data is far greater than that of natural language and audio signal, so the number of embedded tokens is inevitably large. Thus, the distribution of attention scores of tokens becomes smooth. In other words, we face the problem that ViTs cannot attend locally to important visual tokens. The above two main problems cause highly redundant attentions that cannot focus on a target class. This redundant attention makes it easy for ViT to normally concentrate on the background and not capture the shape of the target class well~(see Fig.~\ref{fig:quan}).
    
    
    This paper presents two solutions to effectively improve the locality inductive bias of ViT for learning small-size datasets from scratch. First, we propose Shifted Patch Tokenization (SPT) to further utilize spatial relations between neighboring pixels in the tokenization process. The idea of SPT was derived from Temporal Shift Module (TSM)~\cite{TSM}. TSM is effective temporal modeling which shifts some temporal channels of features. Inspired by this, we propose effective spatial modeling that tokenizes spatially shifted images together with the input image.
    SPT can give a wider receptive field to ViT than standard tokenization. This has the effect of increasing the locality inductive bias by embedding more spatial information in each visual token. Second, we propose Locality Self-Attention (LSA), which allows ViT to attend locally. LSA mitigates the smoothing phenomenon of attention score distribution by excluding self-tokens and by applying learnable temperature to the softmax function.
    LSA induces attention to work locally by forcing each token to focus more on tokens with large relation to itself.
    Note that the proposed SPT and LSA can be easily applied to various ViTs in the form of add-on modules without structural changes and can effectively improve performance~(see Fig.~\ref{fig:main} and Table~\ref{table:ablationTotal}).
    
    
    Our experiments show that the proposed method improves the performance of various ViTs both qualitatively and quantitatively. First, Fig.~\ref{fig:quan} illustrates that when SPT and LSA are applied to the ViTs, object shapes are better captured. From a quantitative aspect, SPT and LSA improve image classification performance. For example, in the experiment on Tiny-ImageNet, the classification accuracy is improved by an average of $2.96$\%, and a maximum of $4.08$\%~(see Table~\ref{table:small_pool}). Also, SPT and LSA improve the performance of ViTs up to $1.06\%$ in the mid-size dataset such as ImageNet~(see Table~\ref{table:ImgaeNet}). The main contribution points of this paper are as follows:
    \begin{itemize}[noitemsep,topsep=0pt]
    \setlength\itemsep{0em}
        \item To sufficiently embed spatial information between neighboring pixels, we propose new tokenization based on spatial feature shifting. The proposed tokenization can give a wider receptive field to visual tokens. This dramatically improves the performance of the ViTs.
        
        \item We propose a locality attention mechanism to solve or attenuate the smoothing problem of the attention score distribution. This mechanism significantly improves the performance of ViTs with only a small parameter increase and the addition of simple operations. 
    \end{itemize}

\input{figures/FigArchi}
\section{RELATED WORK}
\label{sec2:related}
    Recently, several data-efficient ViTs have been proposed to alleviate the dependence of ViT on large-size datasets. These ViTs can learn mid-size datasets from scratch. 
    For example, DeiT~\cite{vit:deit} improved the efficiency of ViTs by employing data augmentations and regularizations and realized knowledge distillation by introducing the distillation token concept. T2T~\cite{vit:T2T} used a tokenization method that flattened overlapping patches and applied a transformer. This makes it possible to learn local structure information around a token. PiT~\cite{vit:pit} produced various receptive fields through spatial dimension reduction based on the pooling structure of a convolutional layer. CvT~\cite{vit:CvT} replaced both linear projection and multi-layer perceptron with convolutional layers. Also, like PiT, CvT generated various receptive fields only with a convolutional layer. 
    Swin Transformer~\cite{vit:swin} presented an efficient hierarchical transformer that gradually reduces the number of tokens through patch merging while using attention calculated in non-overlapping local windows. CaiT~\cite{vit:CaiT} employed LayerScale, which converges well even in training ViTs with a large depth. In addition, the transformer layer of the CaiT is divided into a patch-attention layer and a class-attention layer, which is effective for class embedding.
    
    However, the ViT for small-size datasets has not been reported yet. Therefore, this paper proposes tokenization using more spatial information and also a high-performance attention mechanism, which allows ViTs to effectively learn small-size datasets from scratch.
    
    
    
\section{PROPOSED METHOD}
\label{sec3:method}
    This section specifically describes two key ideas for increasing the locality inductive bias of ViTs: SPT and LSA. First, Fig.~\ref{fig:archi_a} depicts the concept of SPT. SPT spatially shifts an input image in several directions and concatenates them with the input image. Fig.~\ref{fig:archi_a} is an example of shifting in four diagonal directions. Next, patch partitioning is applied like standard ViTs. Then, for embedding into visual tokens, three processes are sequentially performed: patch flattening, layer normalization~\cite{norm:LayerNorm}, and linear projection. As a result, SPT can embed more spatial information into visual tokens and increase the locality inductive bias of ViTs.
    
    Fig.~\ref{fig:archi_b} explains the second idea, LSA. In general, a softmax function can control the smoothness of the output distribution through temperature scaling~\cite{softmax}. LSA primarily sharpens the distribution of attention scores by learning the temperature parameters of the softmax function. Additionally, the self-token relation is removed by applying the so-called diagonal masking, which forcibly suppresses the diagonal components of the similarity matrix computed by Query and Key. This masking relatively increases the attention scores between different tokens, making the distribution of attention scores sharper. As a result, LSA increases the locality inductive bias by making ViT's attention locally focused.
    
    
    \subsection{Preliminary}
        \label{sec3.1:preliminary}
        Before a detailed description of the proposed SPT and LSA, this section briefly reviews the tokenization and formulation of the self-attention mechanism of standard ViT~\cite{vit:vit}.
        
        
        Let $\bm{\mathrm{x}} \in \mathbb{R}^{H \times W \times C}$ be an input image. Here, $H$, $W$, and $C$ indicate the height, width, and channel of the image, respectively.
        First, ViT divides the input image into non-overlapping patches and flatten the patches to obtain a sequence of vectors. This process is formulated as Eq.~\ref{eq:1}:
        \begin{equation}\label{eq:1}
             \mathcal{P}(\bm{\mathrm{x}})=[\bm{\mathrm{x}}_p^1;\bm{\mathrm{x}}_p^2;\dots;\bm{\mathrm{x}}_p^N]
        \end{equation}
        where $ \bm{\mathrm{x}}_p^i \in \mathbb{R}^{P^2 \cdot C}$ represents the $i$-th flattened vector. $P$ and $N=HW/P^2$ stand for the patch size and the number of patches, respectively.
        
        Next, we obtain patch embeddings by linearly projecting each vector into the space of the hidden dimension of the transformer encoder. Each patch embedding corresponds to a visual token input to the transformer encoder, so this series of processes is called tokenization, i.e., $\mathcal{T}$. This is defined by:
        \begin{equation}\label{eq:2}
             \mathcal{T}(\bm{\mathrm{x}})=\mathcal{P}(\bm{\mathrm{x}})\bm{E}_t
        \end{equation}
        where $\bm{E}_t \in \mathbb{R}^{(P^2 \cdot C) \times d}$ is the learnable linear projection for tokens, and $d$ is the hidden dimension of the transformer encoder. 
        
        Note that the receptive fields of visual tokens in ViT are determined by tokenization. In the transformer encoder running after the tokenization step, the number of visual tokens does not change, so the receptive field cannot be adjusted there.
        and the tokenization (Eq.~\ref{eq:2}) of standard ViT is the same as the operation of the non-overlapping convolutional layer with the same size of kernel and stride. So, the receptive field size of visual tokens can be calculated by the following equation given in \cite{receptive}:
        \begin{equation}\label{eq:3}
            r_{token}= r_{trans} \cdot j+(k-j)
        \end{equation}
        where $r_{token}$ and $r_{trans}$ stand for the receptive field sizes of tokenization and transformer encoder, respectively. $j$ and $k$ are the stride and kernel size of the convolutional layer, respectively. As mentioned earlier, the receptive field is not adjusted in the transformer encoder, so $r_{trans}=1$. Thus, $r_{token}$ is the same as the kernel size. Here, the kernel size is the patch size of ViT.
        
        At this time, let's investigate whether $r_{token}$ is of sufficient size. For instance, we compare $r_{token}$ with the receptive field size of the last feature of ResNet50 when training on the ImageNet dataset consisting of images of $224\times224$. The patch size of standard ViT is 16, so $r_{token}$ of visual tokens is also 16. On the other hand, the receptive field size of the ResNet50 feature amounts to 483~\cite{receptive}. As a result, the visual tokens of ViTs have a receptive field size that is about 30 times smaller than that of the ResNet50 feature. We interpret this small receptive field of tokenization as a major factor in the lack of local inductive bias. Therefore, Sec.~\ref{sec3.2:SPT} proposes the SPT to leverage rich spatial information by increasing the receptive field of tokenization.

    
        Meanwhile, the self-attention mechanism of general ViTs operates as follows. First, a learnable linear projection is applied to each token to obtain Query, Key, and Value. Next, calculate the similarity matrix, that is, $\mathrm{R} \in \mathbb{R}^{(N+1) \times (N+1)}$, indicating the semantic relation between tokens through the dot product operation of Query and Key. The diagonal components of $\mathrm{R}$ represent self-token relations, and the off-diagonal components represent inter-token relations:
        \begin{equation}\label{eq:4}
            \mathrm{R}(\bm{\mathrm{x}})=\bm{\mathrm{x}}\bm{E}_q{(\bm{\mathrm{x}}\bm{E}_k)}^\intercal
        \end{equation}
        Here, $\bm{E}_{q} \in \mathbb{R}^{d \times d_q}, \bm{E}_{k} \in \mathbb{R}^{d \times d_k}$ indicate learnable linear projections for Query and Key, respectively. And, $d_q$ and $d_k$ are the dimensions of Query and Key, respectively. Next, $\mathrm{R}$ is divided by the square root of the Key dimension, and then the softmax function is applied to obtain the attention score matrix. Finally, calculate the self-attention, defined by the dot product of the attention score matrix and Value, as in Eq.~\ref{eq:5}:
        \begin{equation}\label{eq:5}
            \mathrm{SA}(\bm{\mathrm{x}})=\mathrm{softmax}(\mathrm{R}/\sqrt{d_k})\bm{\mathrm{x}}\bm{E}_v
        \end{equation}
        where $\bm{E}_{v} \in \mathbb{R}^{d \times d_v}$ is a learnable linear projection of Value, and $d_v$ is the Value dimension.
        
        Eq.~\ref{eq:5} was designed so that the attentions of tokens with large relations get large. However, due to the following two causes, attentions of standard ViT tend to be similar to each other regardless of relations.
        The first cause is as follows: Since Query ($\bm{\mathrm{x}}\bm{E}_q$) and Key ($\bm{\mathrm{x}}\bm{E}_k$) is linearly projected from the same input tokens, token vectors belonging to Query and Key tend to have similar sizes. Eq.~\ref{eq:4} shows that $\mathrm{R}$ is the dot product of Query and Key. So, self-token relations which are dot products of similar vectors are usually larger than inter-token relations. Therefore, the softmax function of Eq.~\ref{eq:5} gives relatively high scores to self-token relations and small scores to inter-token relations. The second cause is as follows: The reason why $\mathrm{R}$ is divided by $\sqrt{d_{k}}$ in Eq.~\ref{eq:5} is to prevent the softmax function from having a small gradient. However, $\sqrt{d_{k}}$ can rather act as a high temperature of the softmax function and cause smoothing of the attention score distribution~\cite{softmax}. Our experiment proves that the attention scores smoothed due to high temperature degrade the performance of ViT. For example, take a look at Table~\ref{table:VariousTemp} that shows the top-1 accuracy of standard ViT on the small-size datasets, i.e., CIFAR100 and Tiny-ImageNet. Here, we can observe the best performance when the temperature of softmax is less than $\sqrt{d_{k}}$. Sec.~\ref{sec3.3:LSA} proposes the LSA for improving the performance of ViTs by solving the smoothing problem of the attention score distribution.
    
    \input{figures/FigT}
    \input{figures/FigKLD}
    
    \subsection{Shifted Patch Tokenization}
        \label{sec3.2:SPT}
        This section first describes the overall formulation of SPT~(Sec.~\ref{sec3.2.1:formula}) and applies the proposed SPT to the patch embedding layer and the pooling layer, i.e., two main tokenizations for ViTs~(Sec.~\ref{sec3.2.2:PE} and Sec.~\ref{sec3.2.3:Pool}).
        
        \subsubsection{Formulation}
        \label{sec3.2.1:formula}
        First, each input image is spatially shifted by half the patch size in four diagonal directions, that is, left-up, right-up, left-down, and right-down. In this paper, this shifting strategy is named $\mathcal{S}$ for convenience, and the SPT of all experiments follows $\mathcal{S}$. Of course, various shifting strategies other than $\mathcal{S}$ are available, and they are dealt with in the supplementary. Next, the shifted features are cropped to the same size as the input image and then concatenated with the input. Then, the concatenated features are divided into non-overlapping patches and the patches are flattened as in Eq.~\ref{eq:1}. Next, visual tokens are obtained through layer normalization ($\mathrm{LN}$) and linear projection. The whole process is formulated as Eq.~\ref{eq:6}:
        \begin{equation}\label{eq:6}
            \mathrm{S}(\bm{\mathrm{x}})=\mathrm{LN}(\mathcal{P}([\bm{\mathrm{x}} \, \bm{\mathrm{s}}^1 \, \bm{\mathrm{s}}^2 \, \dots \, \bm{\mathrm{s}}^{N_{\mathcal{S}}}]))\bm{E}_{\mathcal{S}}
        \end{equation}
        Here, $\bm{\mathrm{s}}^i \in \mathbb{R}^{H \times W \times C}$ represents the $i$-th shifted image according to $\mathcal{S}$ and $\bm{E}_{\mathcal{S}} \in \mathbb{R}^{(P^2 \cdot C \cdot (N_s+1) \times d_{\mathcal{S}})}$ indicates a learnable linear projection. Also, $d_{\mathcal{S}}$ represents the hidden dimension of the transformer encoder, and $N_{\mathcal{S}}$ represents the number of images shifted by $\mathcal{S}$.
        
        \subsubsection{Patch Embedding Layer}
        \label{sec3.2.2:PE}
        This section describes how to use SPT as a patch embedding layer. We concatenate a class token to visual tokens and then add positional embedding. Here the class token is the token with representation information of the entire image, and the positional embedding gives positional information to the visual tokens. If a class token is not used, only positional embedding is added to the output of SPT. How to apply the SPT to the patch embedding layer is formulated as follows:
        \begin{equation}\label{eq:7}
          \mathrm{S}_{pe}(\bm{\mathrm{x}})=
          \begin{cases}
          [\bm{\mathrm{x}}_{cls};\mathrm{S}(\bm{\mathrm{x}})]+\bm{E}_{pos} & \mathrm{if} \, \bm{\mathrm{x}}_{cls} \, \mathrm{exist} \\
          \mathrm{S}(\bm{\mathrm{x}})+\bm{E}_{pos} & \mathrm{otherwise} \\
          \end{cases}
        \end{equation}
        where $\bm{\mathrm{x}_{cls}} \in \mathbb{R}^{d_{\mathcal{S}}}$ is a class token and $\bm{E}_{pos} \in \mathbb{R}^{(N+1) \times d_{\mathcal{S}}}$ is the learnable positional embedding. Also, $N$ is the number of embedded tokens in Eq.~\ref{eq:6}.
        
        \subsubsection{Pooling Layer}
        \label{sec3.2.3:Pool}
        Tokenization is the process of embedding 3D-tensor features into 2D-matrix features. For example, it embeds $\bm{\mathrm{x}} \in \mathbb{R}^{H \times W \times C}$ into $\bm{\mathrm{y}}=\mathcal{T}(\bm{\mathrm{x}}) \in \mathbb{R}^{N \times d}$. Since $N=HW/P^2$, the spatial size of the 3D feature is reduced by $P^2$ through the tokenization process. So, if tokenization is used as a pooling layer, the number of visual tokens can be reduced.
        Therefore, we propose to use SPT as a pooling layer as follows: First, class tokens and visual tokens are separated, and visual tokens in the form of 2D-matrix are reshaped into 3D-tensor with spatial structure, i.e., $\mathcal{R} : \mathbb{R}^{N \times d} \rightarrow \mathbb{R}^{(H/P) \times (W/P) \times d}$. Then, if the SPT of Eq.~\ref{eq:6} is applied, new visual tokens with a reduced number of tokens are embedded. Finally, the linearly projected class token is connected with the embedded visual tokens. If there is no class token, only $\mathcal{R}$ is applied before the output of SPT. The whole process is formulated as Eq.~\ref{eq:8}:
        \begin{equation}\label{eq:8}
          \mathrm{S}_{pool}(\bm{\mathrm{y}})=
          \begin{cases}
            [\bm{\mathrm{x}}_{cls}\bm{E}_{cls};\mathrm{S}(\mathcal{R}(\bm{\mathrm{y}}))] & \mathrm{if} \, \bm{\mathrm{x}}_{cls} \, \mathrm{exist} \\ 
            \mathrm{S}(\mathcal{R}(\bm{\mathrm{y}})) & \mathrm{otherwise}
          \end{cases}
        \end{equation}
        where $\bm{E}_{cls} \in \mathbb{R}^{d \times d_{\mathcal{S}}^{'}}$ is a learnable linear projection. In addition, $d_{\mathcal{S}}^{'}$ is the hidden dimension of the next stage. 
        As a result, SPT embeds rich spatial information into visual tokens by increasing the receptive field of tokenization as much as spatially shifted. 
        
    
    \subsection{Locality Self-Attention Mechanism}    
        \label{sec3.3:LSA}
        \input{tables/VariousTemp}

        This section describes the LSA. The core of LSA is the diagonal masking (Sec.~\ref{sec3.3.1:masking}) and the learnable temperature scaling (Sec.~\ref{sec3.3.2:temp}).
        
        \subsubsection{Diagonal Masking}
        \label{sec3.3.1:masking}
        Diagonal masking plays a role in giving larger scores to inter-token relations by fundamentally excluding self-token relations from the softmax operation. Specifically, diagonal masking forces $-\infty$ on diagonal components of $\mathrm{R}$ of Eq.~\ref{eq:4}. This makes ViT's attention more focused on other tokens rather than attending to its own tokens. The proposed diagonal masking is defined by:
         \begin{equation}\label{eq:9}
            \mathrm{R}_{i,j}^M(\bm{\mathrm{x}})=
            \begin{cases}
                \mathrm{R}_{i,j}(\bm{\mathrm{x}}) & (i\not=j)\\
                -\infty & (i=j)\\
            \end{cases}
        \end{equation}
        where $\mathrm{R}_{i,j}^M$ indicates each component of the masked similarity matrix.

        \subsubsection{Learnable Temperature Scaling}
        \label{sec3.3.2:temp}
        The second technique for LSA is the learnable temperature scaling, which allows ViT to determine the softmax temperature by itself during the learning process. Fig.~\ref{fig:temp} shows the average learned temperature according to depth when the softmax temperature is used as the learnable parameter in Eq.~\ref{eq:5}. Note that the average learned temperature is lower than the constant temperature of standard ViT. In general, the low temperature of softmax sharpens the score distribution. Therefore, the learnable temperature scaling sharpens the distribution of attention scores. Based on Eq.~\ref{eq:5}, the LSA with both diagonal masking and learnable temperature scaling applied is defined by:
        \begin{equation}\label{eq:10}
            \mathrm{L}(\bm{\mathrm{x}})=\mathrm{softmax}(\mathrm{R^{M}(\bm{\mathrm{x}})}/{\mathrm{\tau}})\bm{\mathrm{x}}\bm{E}_v
        \end{equation}
        where $\mathrm{\tau}$ is the learnable temperature. 
        
        In other words, LSA solves the smoothing problem of the attention score distribution. Fig.~\ref{fig:kld} shows the depth-wise averages of total Kullback-Leibler divergence ($D_{KL}^{total}$) for all heads. Here, $\mathrm{T}$ and $\mathrm{M}$ mean that only learnable temperature scaling and diagonal masking is applied to ViTs, respectively, and $\mathrm{L}$ indicates that the entire LSA is applied to ViTs. The lower the average of $D_{KL}^{total}$, the flatter the attention score distribution. We can find that when LSA is fully applied, the average of $D_{KL}^{total}$ is larger by about $0.5$ than standard ViT, so LSA attenuates the smoothing phenomenon of the attention score distribution.

\input{tables/SmallDataset} 
\section{EXPERIMENT}
\label{sec4:ex}
    This section verifies that the proposed method improves the performance of various ViTs through several experiments. Sec.~\ref{sec4.1:setting} describes the settings of the following experiments. Sec.~\ref{sec4.2:quantitative} quantitatively shows that the proposed method effectively improves various ViTs and reduces the gap with CNNs. Finally, Sec.~\ref{sec4.3:qualitative} demonstrates that the ViTs are qualitatively enhanced by visualizing the attention scores of the final class token.

    \subsection{SETTING}
    \label{sec4.1:setting}
        \subsubsection{Environment and Dataset}
        The proposed method was implemented in Pytorch~\cite{Pytorch}. In the small-size dataset experiment (Table~\ref{table:small_pool}), The details of throughput measurement are as follows: The inputs were Tiny-ImageNet, and the batch size was 128, and the GPU was RTX 2080 Ti.

        For small-size dataset experiments, CIFAR-10, CIFAR-100~\cite{dataset:CIFAR}, Tiny-ImageNet~\cite{dataset:timnet}, and SVHN~\cite{dataset:SVHN} were employed and ImageNet~\cite{dataset:IMNET} was employed for the mid-size dataset experiment.
        
        \subsubsection{Model Configurations}
       In the small dataset experiment, in the case of ViT, the depth was set to 9, the hidden dimension was set to 192, and the number of heads was set to 12. This configuration was determined experimentally. And in the ImageNet experiment, we used the ViT-Tiny suggested by DeiT~\cite{vit:deit}. In the case of PiT, T2T, Swin and CaiT, the configurations of PiT-XS, T2T-14, Swin-T and CaiT-XXS24 presented in the corresponding papers were adopted as they were, respectively.
        The performance of ViT improves as the number of tokens increases, but the computational cost increases quadratically.
        We were able to experimentally observe that it was effective when both the number of visual tokens in ViT without pooling and the number of tokens in the intermediate stage of ViT with pooling are 64, considering this trade-off. Accordingly, we modified the baseline models.
        In small-size dataset experiments, the patch size of the patch embedding layer was set to $8$  and the patch size of ViTs using pooling layers such as Swin and PiT was set to $16$. In the ImageNet dataset experiment, the patch size was set to be the same as that used in each paper.
        Also, the hidden dimension of MLP was set to twice that of the transformer in the small dataset experiment, and the configuration used in each paper was applied in the ImageNet experiment.
    
        
        \subsubsection{Training Regime}
        According to DeiT, various techniques are required to effectively train ViTs. Thus, we applied data augmentations such as CutMix~\cite{aug:CM}, Mixup~\cite{aug:MU}, Auto Augment~\cite{aug:AA}, Repeated Augment~\cite{aug:RA} to all models. In addition, regularization techniques such as label smoothing~\cite{LabelSmoothing}, stochastic depth~\cite{StochasticDepth}, and random erasing~\cite{aug:RE} were employed.
        Meanwhile, AdamW~\cite{AdamW} was used as the optimizer. Weight decays were set to 0.05, batch size to 128 (however, 256 for ImageNet), and warm-up to 10 (however, 5 for ImageNet). All models were trained for 100 epochs, and cosine learning rate decay was used. In the small-size dataset experiments, the initial learning rate of ViT and CNNs was set to 0.003, and that of the remaining models was set to 0.001. On the other hand, in the ImageNet experiment, the initial learning rate was set to 0.00025 for all models.
  \input{tables/ImageNet}

    \subsection{QUANTITATIVE RESULT}
    \label{sec4.2:quantitative}
        \subsubsection{Image Classification}
        \label{sec4.2.1:classification}
        This section presents the experimental results for small-size datasets and the ImageNet dataset. In the small-size dataset experiment, Throughput, FLOPs, and the number of parameters were measured in Tiny-ImageNet. 
        
        First, Table~\ref{table:small_pool} shows the performance improvement when the proposed method was applied to ViTs. Here, $\mathrm{SL}$ indicates that both SPT and LSA were applied, and $\mathrm{S}_{pool}$ means the SPT applied to the pooling layer. In most cases, the proposed method effectively improved the performance of ViTs, especially in CIFAR100 and Tiny-ImageNet. 
        For example, in CIFAR100, the performance of CaiT and PiT improved by $+3.43\%$ and $+4.01\%$ respectively and in Tiny-ImageNet, the performance of ViT and Swin improved up to $+4.00\%$ and $+4.08\%$ respectively.
        Also note that the performance was greatly improved only with the acceptable overhead of inference latency. In other words, the cost-effectiveness of the proposed method is remarkable. For example, for ViT, T2T, and CaiT, the proposed method causes only latency overhead of $1.12\%$, $1.15\%$, and $1.06\%$ respectively.
        And it can be seen in the case of PiT and Swin that additional performance improvement can be obtained by replacing the pooling layer with $\mathrm{S}_{pool}$. Therefore, we can find that spatial modeling provided by SPT is effective not only for patch embedding but also for the pooling layer.
        Also, This table shows that the proposed method effectively reduces the gap between ViT and CNN on small-size datasets. For example, SL-CaiT achieves the best performance over ResNet and EfficientNet on all datasets except CIFAR10. SL-Swin also offers better throughput while providing performance comparable to CNNs.

        \input{tables/AblationLSA}

        \input{tables/AblationTotal}

        Table~\ref{table:ImgaeNet} shows performance when training a mid-size dataset ImageNet from scratch. In ViT, SPT was applied only to patch embeddings, and in PiT and Swin, SPT was applied to both patch embedding and pooling layers. We could observe that the proposed method is sufficiently effective for ImageNet. For example, the performance was improved by the proposed method as much as $+1.60\%$ for ViT, $+1.44\%$ for PiT, and $+1.06\%$ for Swin. As a result, we find that the proposed method noticeably improves the ViTs even on mid-size datasets.

        \subsubsection{Ablation Study}
        \input{figures/FigQuan}
        \label{sec4.2.3:ablation}
        This section describes the ablation study on the proposed method. ViT was used for this experiment.
        
        \paragraph{Elements of LSA}
        Let's look at the effect of learnable temperature scaling and diagonal masking, two key elements of LSA, on overall performance.
        Table~\ref{table:ablationLMSA} shows that learnable temperature scaling and diagonal masking effectively resolves the smoothing phenomenon of attention score distribution (see Fig.~\ref{fig:kld}). For example, learnable temperature scaling and diagonal masking in Tiny-ImageNet improved performance by $+0.88\%$ and $+1.22\%$, respectively. Considering that the LSA applied with both techniques shows a performance improvement of $+1.43\%$, we can claim that the contribution of each is sufficiently large and the two techniques produce a synergy.
        \paragraph{SPT and LSA}
        Table~\ref{table:ablationTotal} shows that SPT and LSA can dramatically improve performance by increasing the locality inductive bias of ViT independently. In particular, in Tiny-ImageNet, SPT and LSA improved performance by $+1.43\%$ and $+3.60\%$, respectively. When both techniques were applied, the performance improvement was $+4.00\%$. This proves the competitiveness and synergy of the two key element technologies.

    \subsection{QUALITATIVE RESULT}
    \label{sec4.3:qualitative}
    Fig.~\ref{fig:quan} visualizes the attention scores of the final class token when SPT and LSA were applied to various ViTs. When the proposed method was applied, we can observe that the object shape is better captured as the attention, which was dispersed in the background, is concentrated on the target class. In particular, this phenomenon is evident in the CaiT of the first row, the T2T of the second row, the ViT of the third row, and the PiT of the last row. Therefore, we can find that the proposed method effectively increases the locality inductive bias and induces the attention of the ViTs to improve.
 

\section{CONCLUSION}
\label{sec5:conclusion}
To train ViT on small-size datasets, this paper presents two novel techniques to increase the locality inductive bias of ViT. First, SPT embeds rich spatial information into visual tokens through specific transformation. Second, LSA induces ViT to attend locally through softmax with learnable parameters. The SPT and LSA can achieve significant performance improvement independently, and they are applicable to any ViTs. Therefore, this study proves that ViT learns small-size datasets from scratch and provides an opportunity for ViT to develop further.


{\small
\bibliographystyle{ieee_fullname}
\bibliography{egbib}
}

\input{supplement}

\end{document}

%% file: figures/FigMain.tex
\begin{figure}[t]
\begin{center}
    \includegraphics[width=1\columnwidth]{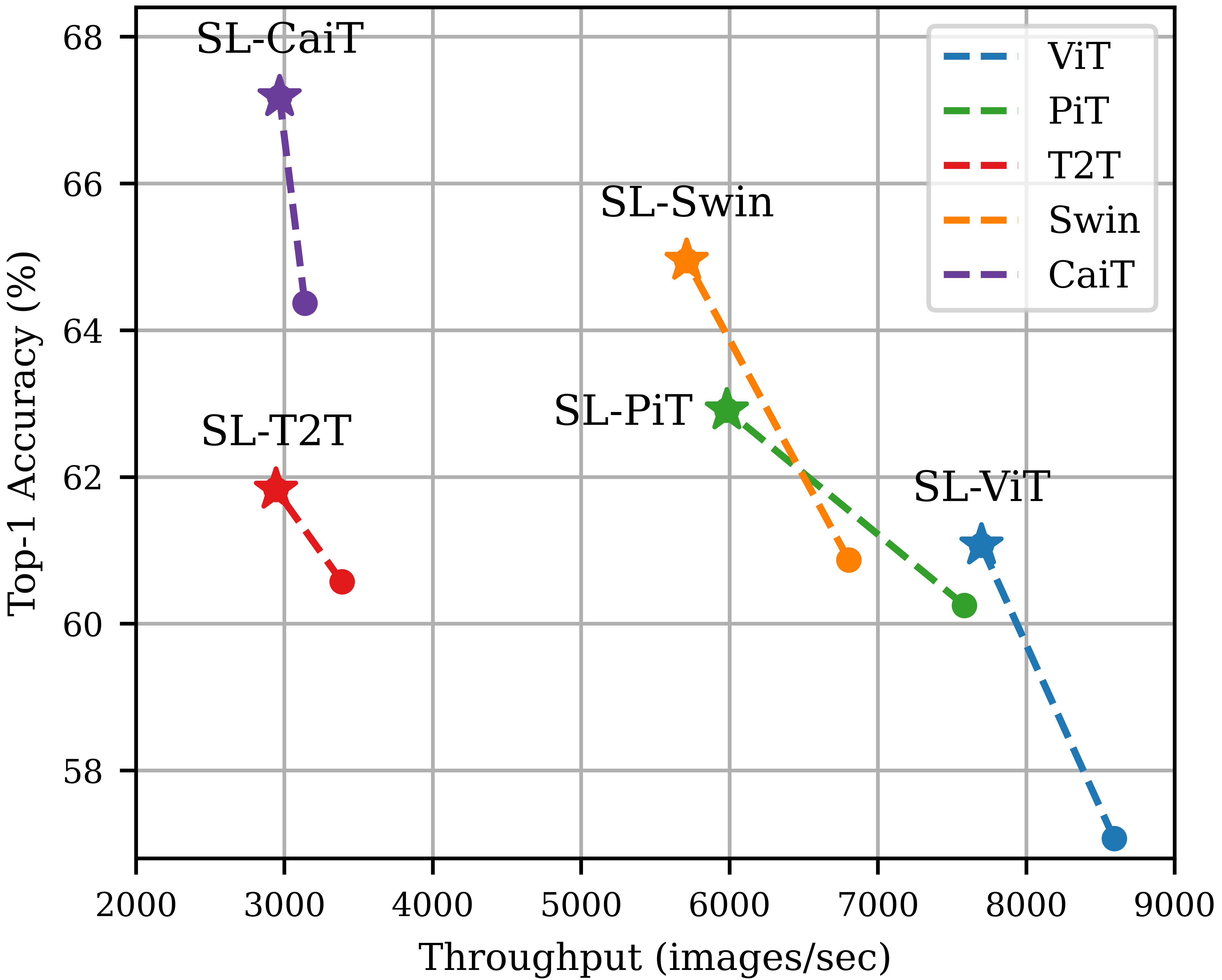}
\end{center}
\caption{Effect of the proposed method on the overall performance when learning Tiny-ImageNet from scratch. Throughput refers to how many images can be processed per unit of time. The stars and dots indicate after and before the proposed method are applied, respectively.}
\label{fig:main}
\end{figure}

%% file: figures/FigArchi.tex
\begin{figure*}[t]
\begin{center}
    \begin{subfigure}[t]{.553\linewidth}
        \centering
        \includegraphics[width=1\linewidth]{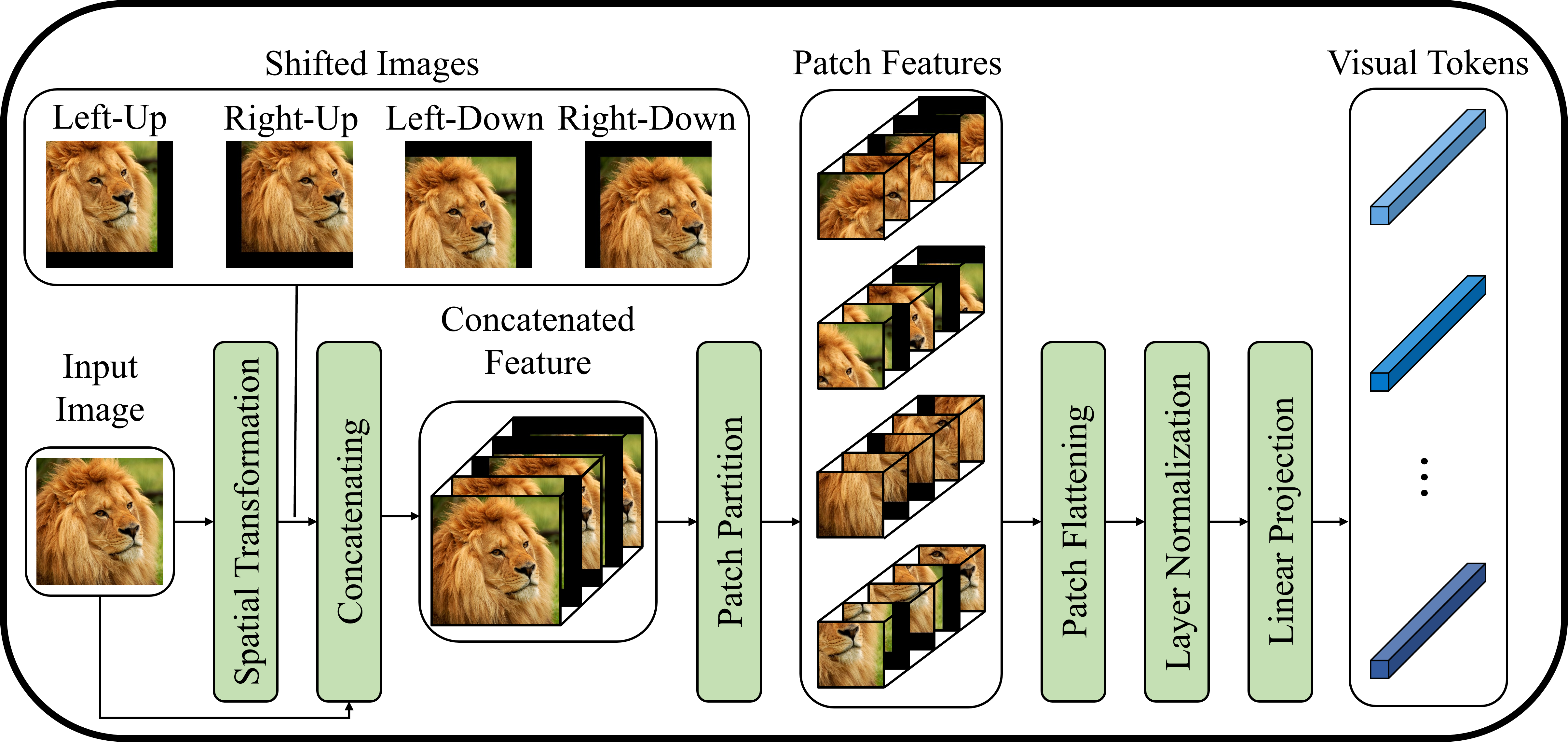}
        \caption{Shifted Patch Tokenization}
        \label{fig:archi_a}
    \end{subfigure}
    \begin{subfigure}[t]{.415\linewidth}
        \centering
        \includegraphics[width=1\linewidth]{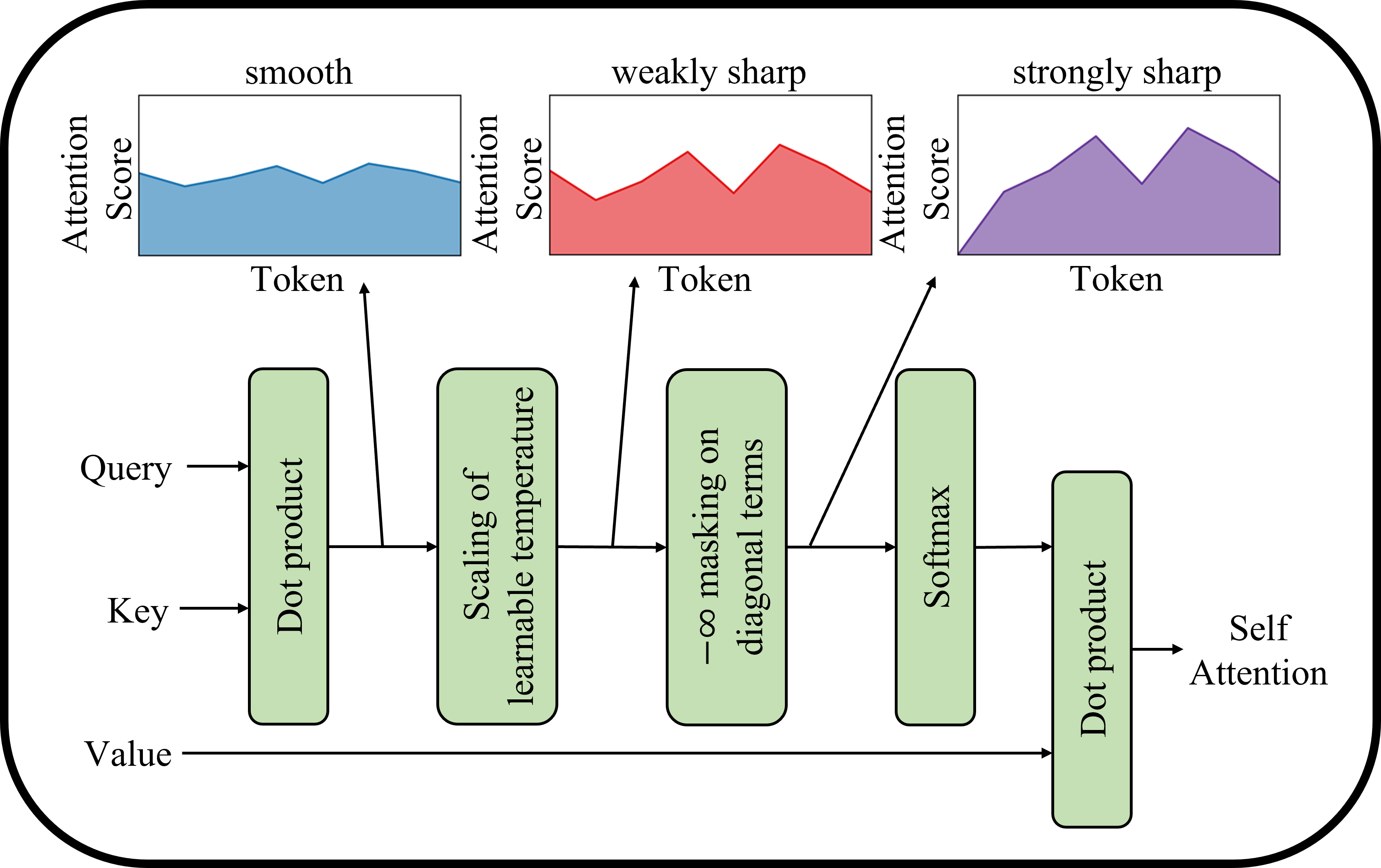}
        \caption{Locality Self-Attention}
        \label{fig:archi_b}
    \end{subfigure}
\end{center}
\caption{Architectures of the proposed SPT and LSA.}
\label{fig:archi}
\end{figure*}

%% file: figures/FigT.tex
\begin{figure}[t]
\begin{center}
    \includegraphics[width=0.95\columnwidth]{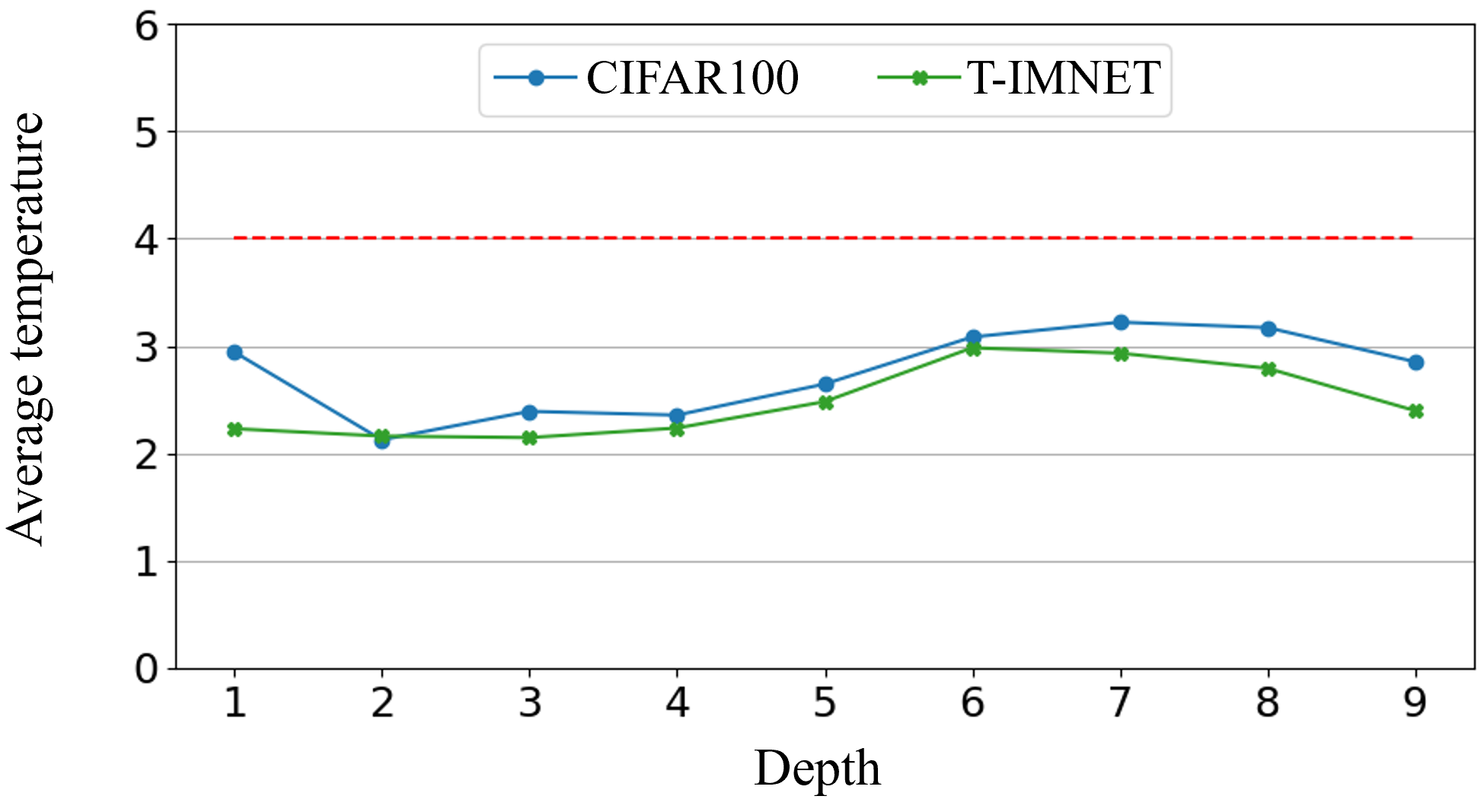}
\end{center}
\caption{The learned temperature according to depth. Here, the red dashed line indicates the temperature of standard ViT.}
\label{fig:temp}
\end{figure}

%% file: figures/FigKLD.tex
\begin{figure}[t]
\begin{center}
    \includegraphics[width=0.95\columnwidth]{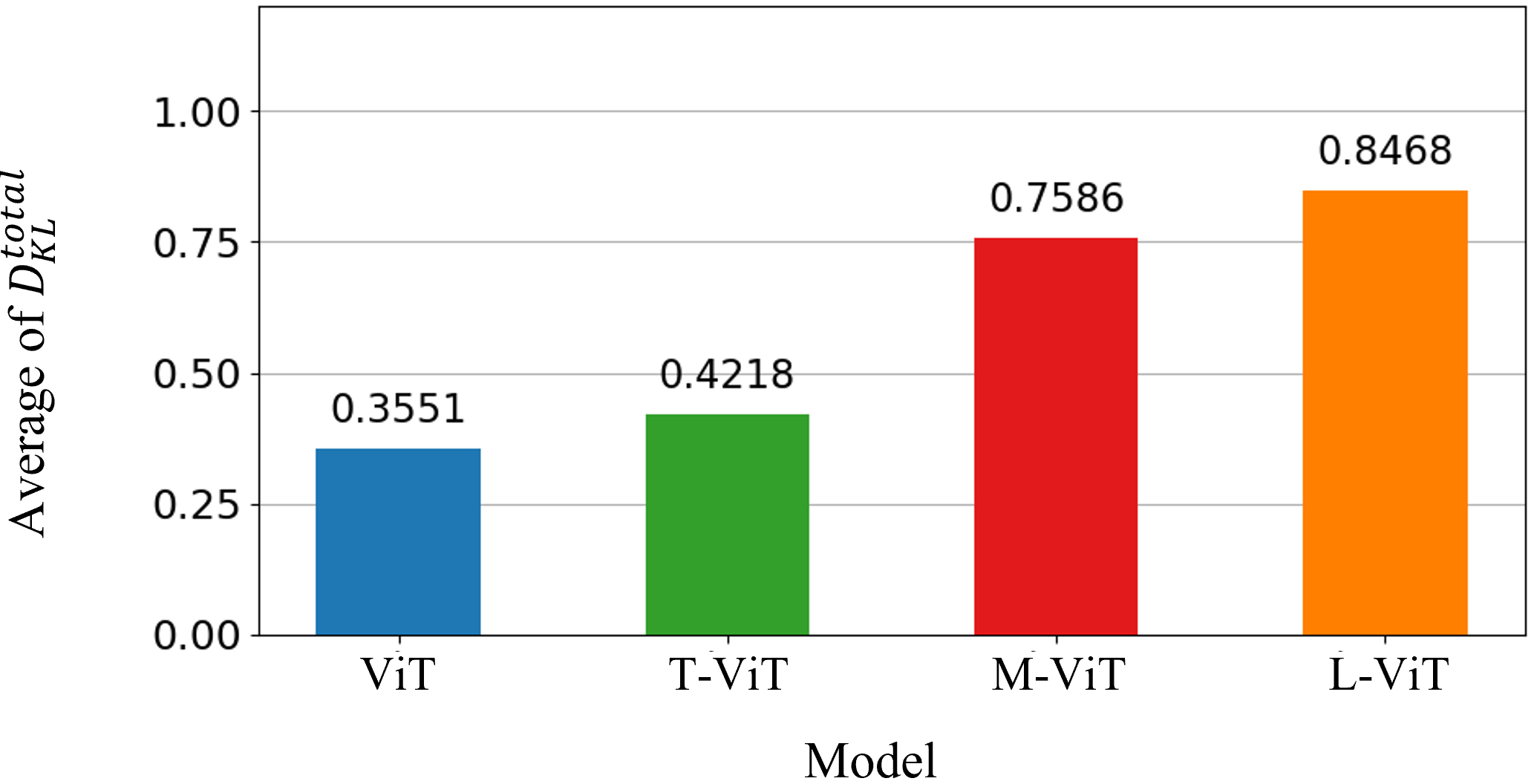}
\end{center}
\caption{Kullback–Leibler Divergence (KLD) of attention score distributions. The average KLDs were measured on Tiny-ImageNet.}
\label{fig:kld}
\end{figure}

%% file: tables/VariousTemp.tex
\begin{table}[t]
\caption{Top-1 accuracy (\%) according to temperatures.}
\begin{center}
\begin{tabular}{ccc}
\toprule
    \multirow{2}{*}{\textbf{TEMPERATURE}}&
    \multicolumn{2}{c}{$\textbf{TOP-1 ACCURACY (\%)}$}\\
    \cmidrule{2-3}
    & \textbf{CIFAR100}& \textbf{T-ImageNet}\\
\midrule
    $\frac{1}{4}\sqrt{d_{k}}$&   $73.70$ & $57.62$ \\
    $\frac{1}{2}\sqrt{d_{k}}$&     $\textbf{74.54}$ & $\textbf{57.65}$ \\  
    $\sqrt{d_{k}}$&       $73.81$ & $57.07$ \\  
    $2\sqrt{d_{k}}$&     $72.77$ & $56.98$ \\  
    $4\sqrt{d_{k}}$&     $71.55$ & $56.43$ \\  
\bottomrule
\end{tabular}
\end{center}
\label{table:VariousTemp}
\end{table}

%% file: tables/SmallDataset.tex
\begin{table*}[t]
\caption{Top-1 accuracy comparison of different models on small-size datasets.}
\begin{center} 
\Large
\resizebox{2\columnwidth}{!}{
\begin{tabular}{cccccccc}
\toprule[1.3pt]
    \multirow{2}{*}{\textbf{MODEL}}&\textbf{THROUGHPUT}&\textbf{FLOPs}&\textbf{PARAMS}&\multirow{2}{*}{\textbf{CIFAR10}}&   \multirow{2}{*}{\textbf{CIFAR100}}&\multirow{2}{*}{\textbf{SVHN}}& \multirow{2}{*}{\textbf{T-ImageNet}}\\
    &\textbf{(images/sec)}&\textbf{(M)}&\textbf{(M)}&&&&\\
\midrule
    ResNet 56  & $4295$ &$506.2$&   $0.9$& $95.70$  & $76.36$  & $97.73$  & $58.77$  \\
    ResNet 110  & $2143$ &$1020.0$&   $1.7$& $\bf96.37$  & $79.86$ & $97.85$  & $62.96$  \\
    EfficientNet B0 & $4078$ &$123.9$&   $3.7$& $94.66$  & $76.04$ & $97.22$  & $66.79$  \\
\midrule
    ViT  & $8593$ &$189.8$&   $2.8$& $93.58$  & $73.81$  & $97.82$  & $57.07$  \\
    SL-ViT  & $7697$ &$199.2$&   $2.9$& $94.53$  & ${76.92}$ & $97.79$  & $61.07$  \\
\midrule
    T2T & $3388$ &$643.0$&  $6.7$& $95.30$  & $77.00$  & $97.90$  & $60.57$  \\
    SL-T2T & $2943$ &$671.4$&  $7.1$& $95.57$  & $77.36$  & $97.91$  & $61.83$  \\
\midrule
    CaiT & $3138$ &$613.8$&  $9.1$& $94.91$  & $76.89$  & $98.13$  & $64.37$  \\
    SL-CaiT & $2967$ &$623.3$&  $9.2$& $95.81$  & $\bf80.32$ & $\bf98.28$  & $\bf67.18$  \\
    
\midrule
    
    PiT & $7583$ &$279.2$&  $7.1$& $94.24$  & $74.99$ & $97.83$  & $60.25$  \\
    SL-PiT w/o $\mathrm{S}_{pool}$  & $6632$ &$280.4$&  $7.1$ & $94.96$  & $77.08$ & $97.94
    $  & $60.31$  \\
    SL-PiT w/ $\mathrm{S}_{pool}$ & $5981$ &$322.9$&  $8.7$& $95.88$  & $79.00$ & $97.93$  & $62.91$  \\
    \midrule
    Swin & $6804$ &$242.3$&  $7.1$& $94.46$  & $76.87$  & $97.72$  & $60.87$  \\
    SL-Swin w/o $\mathrm{S}_{pool}$ & $6384$ &$247.0$&  $7.1$& $95.30$  & $78.13$ & $97.88$  & $62.70$  \\
    SL-Swin w/ $\mathrm{S}_{pool}$ & $5711$ &$284.9$&  $10.2$& $95.93$  & $79.99$ & $97.92$  & $64.95$  \\
    
     
\bottomrule[1.3pt]
\end{tabular}
}
\end{center}%
\label{table:small_pool}
\end{table*}

%% file: tables/ImageNet.tex
\begin{table}[t]
\caption{Top-1 accuracy (\%) of the proposed method on ImageNet dataset.}
\begin{center} 
\begin{tabular}{cc}
\toprule
    \textbf{MODEL}& $\textbf{TOP-1 ACCURACY (\%)}$\\
\midrule
    ViT&                $69.95$\\
    SL-ViT&       $71.55\,(+1.60)$\\
\midrule
    PiT&                $75.58$\\  
    SL-PiT&       $77.02\,(+1.44)$\\  
\midrule
    Swin&               $79.95$\\    
    SL-Swin&      $81.01\,(+1.06)$\\    
\bottomrule
\end{tabular}

\end{center}%
\label{table:ImgaeNet}
\end{table}

%% file: tables/AblationLSA.tex
\begin{table}[t]
\caption{Effect of each component of LSA on performance.}
\begin{center} 
\begin{tabular}{ccc}
\toprule
    \multirow{2}{*}{\textbf{MODEL}}& \multicolumn{2}{c}{$\textbf{TOP-1 ACCURACY (\%)}$}\\
    \cmidrule{2-3}
    & \textbf{CIFAR100}& \textbf{T-ImageNet}\\
\midrule
    ViT&            $73.81$ & $57.07$ \\
    T-ViT&          $74.35$ & $57.95$ \\  
    M-ViT&          $74.34$ & $58.29$ \\  
    L-ViT&       $\textbf{74.87}$ & $\textbf{58.50}$ \\  
\bottomrule
\end{tabular}
\end{center}%
\label{table:ablationLMSA}
\end{table}

%% file: tables/AblationTotal.tex
\begin{table}[t]
\caption{Effect of the proposed SPT (S) and LSA (L) on performance.}
\begin{center}
\begin{tabular}{ccc}
\toprule
    \multirow{2}{*}{\textbf{MODEL}}& \multicolumn{2}{c}{$\textbf{TOP-1 ACCURACY (\%)}$}\\
    \cmidrule{2-3}
    & \textbf{CIFAR100}& \textbf{T-ImageNet}\\
\midrule
    ViT&                $73.81$ & $57.07$ \\
    L-ViT &           $74.87$ & $58.50$ \\  
    S-ViT &            $76.29$ & $60.67$ \\  
    SL-ViT &       $\textbf{76.92}$ & $\textbf{61.07}$ \\   
\bottomrule
\end{tabular}
\end{center}%
\label{table:ablationTotal}
\end{table}

%% file: figures/FigQuan.tex
\begin{figure*}[t]
\begin{center}
    \includegraphics[width=2\columnwidth]{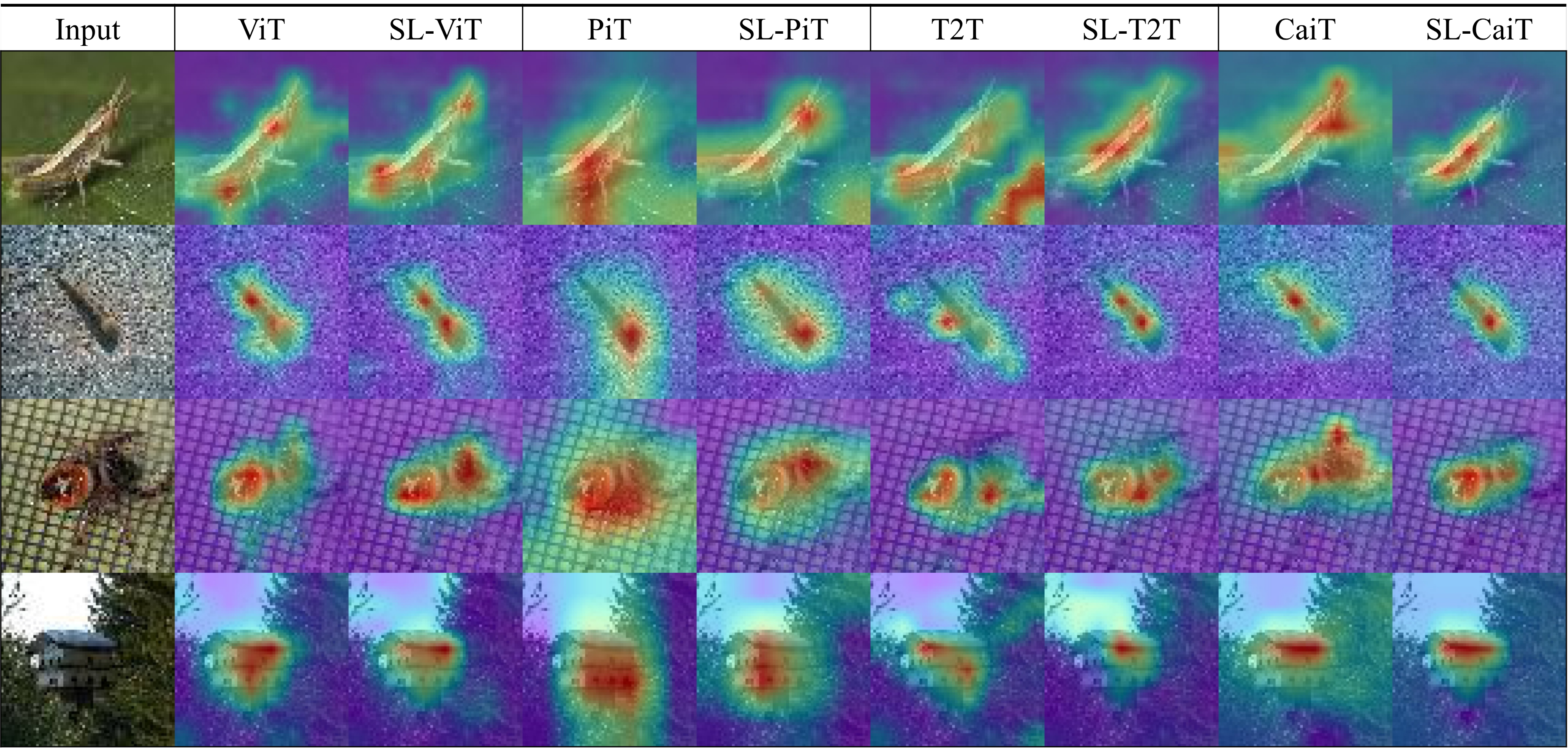}
\end{center}
\caption{Visualization of attention scores of final class tokens.}
\label{fig:quan}
\end{figure*}

%% file: supplement.tex
\newpage
\section*{Supplementary}

This section investigates the various shifting strategies that SPT can employ. Specifically, we explored the shift direction and shift intensity (shift ratio), which have the most impact on performance.

We examined the following three shift directions. The first is the 4 cardinal directions consisting of up, down, left and right directions (Fig.~\ref{fig:directions}(a)). The second is 4 diagonal directions including up-left, up-right, down-left and down-right (Fig.~\ref{fig:directions}(b)). The last is the 8 cardinal directions including all the preceding directions (Fig.~\ref{fig:directions}(c)). Table~\ref{table:shifintDirecitons} shows top-1 accuracy in small-size datasets such as CIFAR-10, CIFAR-100, SVHN, and Tiny-ImageNet for each shift direction. This experiment adopted a model applying SPT to standard ViT. 4 cardinal directions showed the best performance in CIFAR-10 and SVHN. On the other hand, 4 diagonal directions and 8 cardinal directions provided the best performance in CIFAR-100 and Tiny-ImageNet, respectively.
This shows that the shift direction is somewhat dependent on the characteristics of datasets. For example, in CIFAR-10 or CIFAR-100, the target class tends to be in the center of the image, whereas other datasets do not. The location of the target class has some degree of correlation with the shift direction, and the correlation can affect the performance. However, since the performance difference was experimentally marginal, in this paper, the shift direction in the experiment was fixed to 4 diagonal directions.

\input{tables/ShiftDirection}

\input{figures/ShiftDirections}

Next, we look at various shift ratios. The degree of image shifting in SPT is defined as follows: $\mathrm{SHIFT}= P \times r_{shift}$, where $P$ represents the patch size, and $r_{shift}$ represents the shift ratio. Table~\ref{table:ShiftProportions} shows the performance according to shift ratio for CIFAR-100, Tiny-ImageNet, and ImageNet. In this experiment, a model with SPT applied to standard ViT was used, and 4 diagonal directions were adopted. In CIFAR-100 and ImageNet, a ratio of 0.5 was the best, and in Tiny-ImageNet, a ratio of 0.25 was the best. This experimental result shows that the optimal shift ratio also depends on the datasets. Since the relatively most reasonable shift ratio is 0.5 according to our experiment, all the experiments in this paper fixed the shift ratio to 0.5. Note that more various shifting strategies will be available in addition to the methods considered here. The exploration of optimal shifting strategy according to datasets remains as a further work.

\input{tables/ShiftingProportion}




%% file: tables/ShiftDirection.tex
\setcounter{table}{0}
\begin{table}[h]
\caption{Top-1 Accuracy (\%) of Various Shift Directions.}
\begin{center}
\LARGE
\resizebox{1\columnwidth}{!}{
\begin{tabular}{ccccc}
\toprule
    \textbf{DIREC-}&
    \multicolumn{4}{c}{$\textbf{TOP-1 ACCURACY (\%)}$}\\
    \cmidrule{2-5} \textbf{TIONS}
    & \textbf{CIFAR10}& \textbf{CIFAR100}& \textbf{SVHN}& \textbf{T-ImageNet}\\
\midrule
    4 Cardinal&   $\textbf{94.44}$ &          $76.29$ & $\textbf{97.87}$ & $60.35$ \\
    4 Diagonal&    $94.33$ &     $76.29$ &  $97.86$ & $\textbf{60.67}$ \\  
    8 Cardinal&   $94.41$ &      $\textbf{76.40}$ & $97.81$ & $60.57$ \\  
\bottomrule
\end{tabular}
}
\end{center}%
\label{table:shifintDirecitons}
\end{table}

%% file: figures/ShiftDirections.tex
\setcounter{figure}{0}
\begin{figure}[h]
\begin{center}
    \includegraphics[width=1\columnwidth]{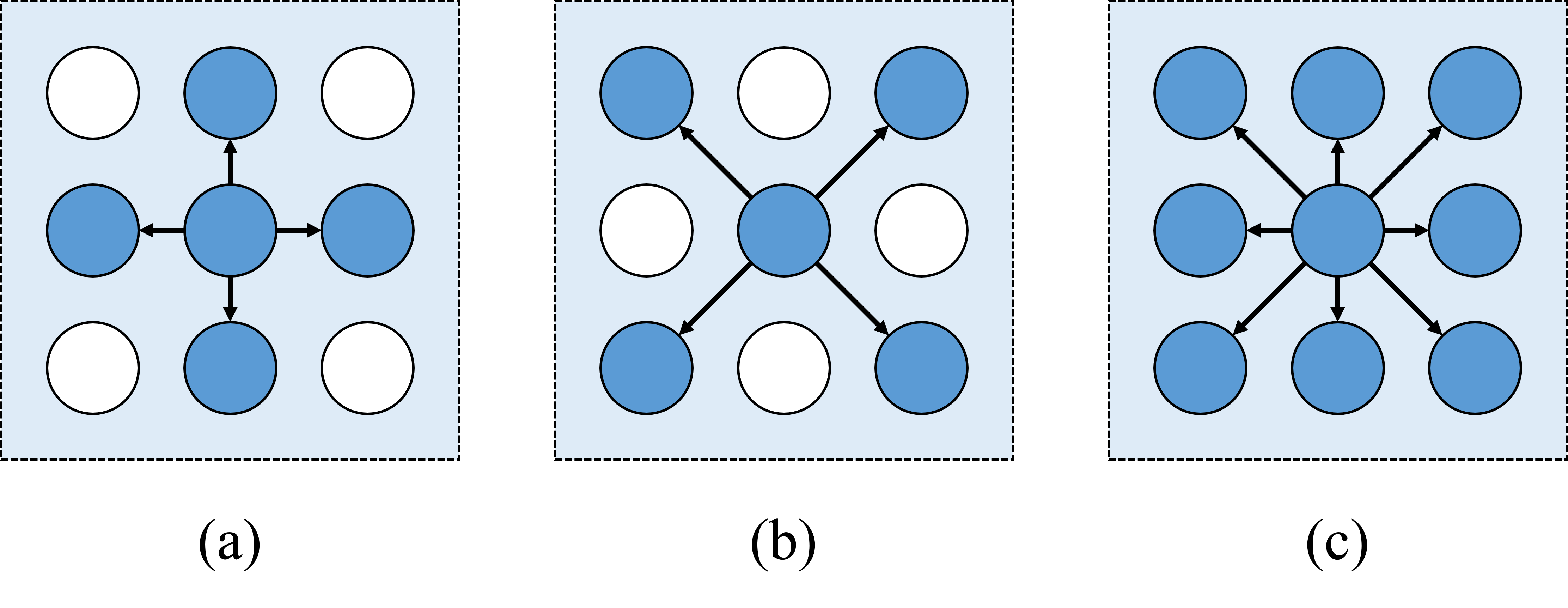}
\end{center}
\caption{Various Shift Directions.}
\label{fig:directions}
\end{figure}

%% file: tables/ShiftingProportion.tex
\begin{table}[h]
\caption{Top-1 Accuracy (\%) of Various Raitos.}
\begin{center}
\small
\begin{tabular}{cccc}
\toprule
    \textbf{SHIFT}&
    \multicolumn{3}{c}{$\textbf{TOP-1 ACCURACY (\%)}$}\\
    \cmidrule{2-4}
    \textbf{RATIO}& 
    \textbf{CIFAR100}& \textbf{T-ImageNet}&
    \textbf{ImageNet}
    \\
\midrule
    $0.125$&   - & $60.63$ & -\\
    $0.25$&   $76.24$ & $\textbf{61.01}$ & $70.65$\\
    $0.5$&     $\textbf{76.29}$ & $60.78$ & $\textbf{70.83}$\\  
    $0.75$&       $75.73$ & $60.18$ & $70.57$\\
    $1.00$&     $74.63$ & $59.35$ & -\\  
\bottomrule
\end{tabular}
\end{center}%
\label{table:ShiftProportions}
\end{table}